\documentclass{article}

\usepackage{arxiv}

\usepackage[utf8]{inputenc} 
\usepackage[T1]{fontenc}    
\usepackage{hyperref}       
\usepackage{url}            
\usepackage{booktabs}       
\usepackage{amsfonts}       
\usepackage{nicefrac}       
\usepackage{microtype}      
\usepackage{lipsum}		
\usepackage{graphicx}
\usepackage[sort&compress,numbers]{natbib}
\usepackage{doi}
\usepackage{comment}
\usepackage{subcaption}
\usepackage{amsmath}

\title{Graph Transformers for inverse physics: reconstructing flows around arbitrary 2D airfoils}

\author{ \href{https://orcid.org/0000-0002-0895-6766}{\includegraphics[scale=0.06]{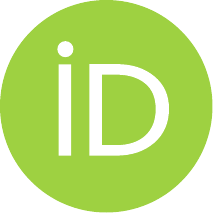}}\hspace{1mm}Gregory Duthé \\
	ETH Zürich\\
	Switzerland \\
	\texttt{duthe@ibk.baug.ethz.ch} \\
	\And
	\href{https://orcid.org/0000-0001-8678-0965}{\includegraphics[scale=0.06]{orcid.pdf}}\hspace{1mm}Imad Abdallah \\
	RTDT Laboratories AG\\
	Switzerland\\
	\texttt{ai@rtdt.ai} \\
    \And
    \href{https://orcid.org/0000-0002-6870-240X}{\includegraphics[scale=0.06]{orcid.pdf}}\hspace{1mm}Eleni Chatzi \\
	ETH Zürich\\
	Switzerland \\
	\texttt{chatzi@ibk.baug.ethz.ch} \\
}

\date{}

\renewcommand{\headeright}{}
\renewcommand{\shorttitle}{}

\hypersetup{
pdftitle={Graph Transformers for inverse physics: reconstructing flows around arbitrary 2D airfoils},
pdfsubject={cs.CE, cs.AI},
pdfauthor={Gregory Duthé, Imad Abdallah, Eleni Chatzi},
pdfkeywords={Graph transformers, Inverse problems, Flow reconstruction},
}

\begin{document}
\maketitle

\begin{abstract}
We introduce a Graph Transformer framework that serves as a general inverse physics engine on meshes, demonstrated through the challenging task of reconstructing aerodynamic flow fields from sparse surface measurements. While deep learning has shown promising results in forward physics simulation, inverse problems remain particularly challenging due to their ill-posed nature and the difficulty of propagating information from limited boundary observations. Our approach addresses these challenges by combining the geometric expressiveness of message-passing neural networks with the global reasoning of Transformers, enabling efficient learning of inverse mappings from boundary conditions to complete states. We evaluate this framework on a comprehensive dataset of steady-state RANS simulations around diverse airfoil geometries, where the task is to reconstruct full pressure and velocity fields from surface pressure measurements alone. The architecture achieves high reconstruction accuracy while maintaining fast inference times. We conduct experiments and provide insights into the relative importance of local geometric processing and global attention mechanisms in mesh-based inverse problems. We also find that the framework is robust to reduced sensor coverage. These results suggest that Graph Transformers can serve as effective inverse physics engines across a broader range of applications where complete system states must be reconstructed from limited boundary observations.
\end{abstract}


\section{Introduction}
Inverse physics problems involve reconstructing the complete state of a system that has produced a set observations - a fundamental challenge that appears across scientific domains~\cite{Tanaka2021}. While traditional numerical physics simulators, such as computational fluid dynamics (CFD) or finite element analysis (FEA) solvers, excel at computing system behavior from known initial conditions, many engineering applications require solving the inverse problem. However, inferring complete system states from limited measurements is challenging~\cite{o1986statistical}. Inverse physics problems are often characterized by ill-posedness~\cite{hadamard1923lectures} and non-uniqueness, where multiple solutions may exist for a given set of observations. Moreover, the derived system state may be extremely sensitive, with small changes in the measured observations leading to large changes in the solution.

One such inverse problem that is of particular interest to engineers dealing with aerodynamic or hydrodynamic applications is the reconstruction of flow fields around immersed structures. At first glance, this may seem extremely difficult, as even the forward modeling of turbulent flows is inherently complex. Yet, some biological systems demonstrate remarkable intuitive ability in flow sensing and reconstruction. Fish and marine mammals, for instance, use their sensory organs to detect minute flow perturbations, allowing for precise navigation and prey detection even in turbulent environments~\cite{dehnhardt2001hydrodynamic, jiang2016investigation}. This insight has motivated the development of artificial sensing systems that aim to replicate such capabilities~\cite{zhao2024comprehensive}. Moreover, recent advances in micro-electromechanical systems (MEMS) have made low-cost, distributed flow sensing increasingly practical and cost-efficient. The availability of these new sensors enables unprecedented spatial coverage for surface measurements in the field~\cite{barber2022development, vandervoort2023development, polonelli2023instrumentation}, providing rich information on surrounding flow. With more detailed boundary conditions and flow features captured by these denser sensor arrays, the potential for high-fidelity flow reconstruction only increases. However, a robust methodology for accurate full-flow field estimation remains an open challenge. High-fidelity flow reconstruction capabilities could have a significant impact on a number of engineering applications: from enhancing the navigation of underwater~\cite{zhou2015bio, yen2018controlling} or aerial~\cite{oettershagen2019real} autonomous vehicles, to enabling real-time condition monitoring of wind turbines~\cite{duthe2021modeling, abdallah2022identifying, scharer2024towards} and aircraft~\cite{raab2021dynamic}, to predicting air quality in urban settings~\cite{apte2024high}.

On the other hand, recent advances in machine learning, particularly in geometric deep learning~\cite{bronstein2021geometric}, have opened new possibilities for tackling forward physics problems with unprecedented accuracy and computational efficiency~\cite{sanchez2018graph, sanchez2020learning}. Geometric deep learning extends traditional neural networks to non-Euclidean domains such as graphs, achieving remarkable success by decomposing complex problems into local interactions and systematically incorporating the inherent geometric priors of the studied physical systems - for instance, meshes~\cite{pfaff2020learning} or molecular structures in chemical systems~\cite{jumper2021highly}. We aim to leverage the advantages of such graph-based learning methodologies to solve inverse physics problems. In particular, the present work demonstrates how Graph Transformers (GTs)~\cite{dwivedi2020generalization, ying2021transformers}, an emerging class of geometric deep learning methods, can be applied to reconstruct the averaged flows around arbitrary airfoils in various high-Reynolds inflow configurations.

The remainder of this paper is organized as follows. Section~\ref{sec:problem} formulates inverse physics problems and flow reconstruction as a graph-based learning framework and discusses the key technical challenges. Section~\ref{sec:background} reviews relevant background and related work across geometric deep learning, flow reconstruction, and discretization-independent PDE operator learning. Section~\ref{sec:dataset} details our dataset generation pipeline, including airfoil selection, meshing strategy, and simulation methodology. Section~\ref{sec:method} introduces the Flow Reconstruction Graph Transformer architecture, describing its core components and design choices. Section~\ref{sec:results} presents experimental results, analyzing architectural variants and evaluating reconstruction capabilities. Section~\ref{sec:discussion} discusses broader implications, limitations, and future directions. Finally, section~\ref{sec:conclusion} concludes with key findings and potential impact.

\section{Problem statement}
\label{sec:problem}

\begin{figure}[h]
  \centering
  \includegraphics[width=0.9\linewidth]{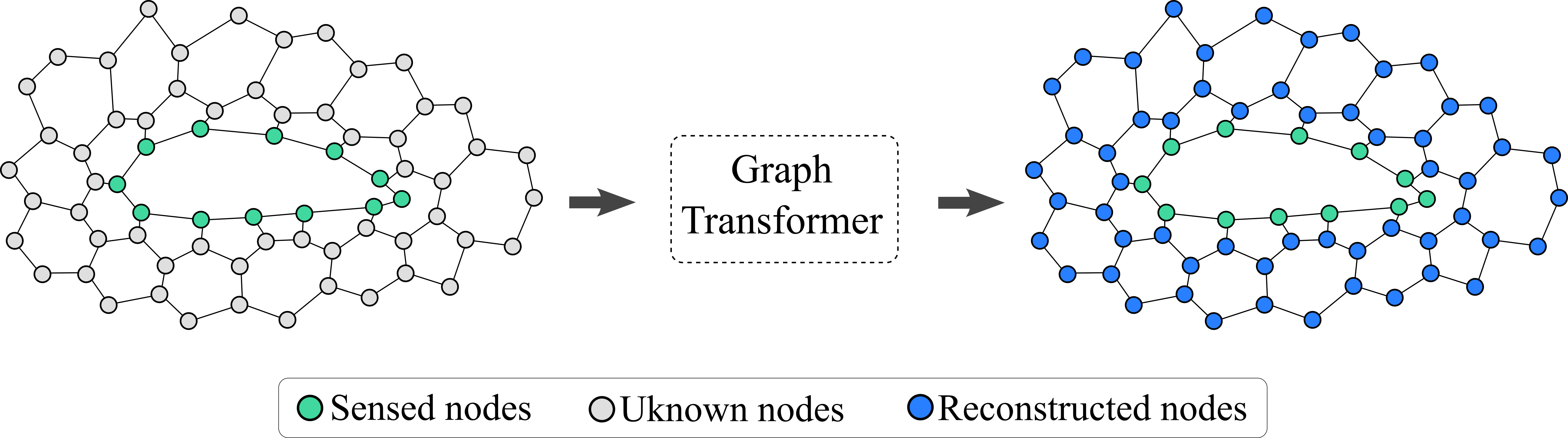}
  \caption{Depiction of an inverse physics problem on a graph. Given a set of measurements on the sensed nodes in our input graph, we aim to inversely reconstruct the response solution at the remaining unmeasured graph nodes. Often, the number of sensed nodes is a fraction of the total nodes.}
  \label{fig:problem_setup}
\end{figure}

\subsection{Graph-based learning for inverse physics}
Adopting the notation of \cite{erichson2020shallow}, an inverse physics problem can be described as follows. Consider a physical system equipped with $p$ distributed sensors that measure some quantity of interest, providing measurements $s \in \mathbb{R}^p$ at discrete locations. These sensors sample from the surrounding physical field $x \in \mathbb{R}^m$ through a measurement operator $H$:
\begin{equation}
s = H(x)
\end{equation}
The goal is to construct an estimate $\hat{x}$ of all variables of the complete field from the available sparse measurements by learning from the training data a function $\mathcal{F}$ that approximates the highly nonlinear inverse measurement operator $G$ such that:
\begin{equation}
\mathcal{F}(s) = \hat{x} \approx x = G(s)
\end{equation}
When the domain can be mapped to a discrete graph structure $\mathcal{G}$, the field reconstruction problem naturally maps to a graph-based learning framework. In this case, a graph $\mathcal{G} = (\mathcal{V}, \mathcal{E})$ represents the physical domain, with $n$ field nodes $\mathcal{V}_{f}$, where quantities are unknown, and $p$ measurement nodes $\mathcal{V}_{p}$, where sensor data is available. The edges $\mathcal{E}$ encode the spatial connectivity and the geometrical relationships between the nodes. Our aim is then to train a data-driven graph operator that estimates the information at the field nodes using only the information contained at the measurement nodes and the topology of the graph.

\subsection{Flow reconstruction around airfoils}
In the context of flow reconstruction around airfoils, this graph framework takes a concrete form, wherein we use CFD meshes as the foundation for the graph representation. The measurement nodes $\mathcal{V}_{m}$ correspond to $p$ pressure sensors distributed along the airfoil surface, providing surface pressure measurements, while the field nodes $\mathcal{V}_{f}$ represent points in the surrounding fluid domain where we aim to reconstruct flow features, such as velocity components and pressure. The mesh cell connectivity naturally defines the edges $\mathcal{E}$, which capture the spatial relationships. This mesh-to-graph conversion preserves the spatial discretization of the original CFD mesh, including refined regions near the airfoil surface. In this setting, the graph operator learns to map from sparse surface measurements to the complete flow field by leveraging both the geometric structure of the problem and the underlying physics. \autoref{fig:problem_setup} illustrates this setup.


\subsection{Challenges in graph-based flow reconstruction}
\label{sec:challenges}
This work extends previous research on graph-based flow reconstruction around airfoils, which utilizes graph neural networks (GNNs)~\cite{duthe2023graph}. We discuss here several key challenges that such an approach, which is based purely on message-passing, may encounter. These points directly motivate some of the architectural choices in our present work.

The graph-based flow reconstruction problem presents several fundamental challenges at the intersection of geometric deep learning and fluid dynamics. The first major hurdle stems from the sheer scale and the connectivity patterns of the aerodynamic CFD meshes that we use as input for our graph-learning setup. These meshes typically contain upwards of $50,000$ nodes with strictly local connectivity patterns. Near the airfoil surface, mesh resolution becomes particularly dense, creating highly unbalanced graph structures. Given that full graphs must be fed to our model and that memory requirements for graph-based learning scales strongly with graph size, the type and the size of the models that can be used is severely limited for standard computational resources (single GPU setup).

A second critical challenge arises from the extreme localization of the input data. Pressure measurements are confined to approximately 1\% of all graph nodes, located on the airfoil surface, yet this information must propagate through multiple dense mesh layers to reach distant regions of the flow field. This is particularly problematic for reconstructing critical flow features like wakes and recirculation zones using surface pressure measurements as this requires effective long-range information transfer. Detached flow is especially challenging in this regard, as the separated region is by definition a zone in which the flow features are disconnected from the surface measurements.

These challenges are further compounded by the architectural limitations of typical graph neural networks, which rely on message-passing. Message-passing neural networks (MPNNs) operate through strictly-local computations, meaning information can only propagate to immediate neighbors for each layer. Consequently, capturing long-range dependencies requires many successive message-passing operations. However, extensively increasing the depth of graph neural networks introduces other fundamental difficulties, such as oversquashing and oversmoothing. The oversquashing phenomenon~\cite{alon2020bottleneck} occurs when information from exponentially growing neighborhoods gets compressed through bottleneck structures, while oversmoothing~\cite{rusch2023survey} causes node features to become indistinguishable after many layers. Moreover, vanishing gradients can make training unstable at extreme depths.

Finally, the underlying physics adds another layer of complexity. High-Reynolds flows exhibit complex nonlinear behavior with turbulent effects creating multi-scale features that require both local and global reasoning. The solution space is also highly sensitive to boundary conditions and geometric variations, making the learning task particularly challenging. This combination of scale, information propagation, architectural, and physical challenges necessitates careful consideration in the design of learning frameworks for large-scale fluid dynamics problems.

\section{Background and related work}
\label{sec:background}
\paragraph{Geometric deep learning}
The field of geometric deep learning~\cite{bronstein2021geometric} is a rapidly evolving domain within the broader machine learning landscape. Geometric deep learning has attracted significant attention in recent years thanks to the notable success in tackling problems that do not live in standard Euclidean spaces~\cite{jumper2021highly, pfaff2020learning}. Geometric deep learning is built on two fundamental principles: symmetry and scale separation. Symmetry (or invariance/equivariance) refers to the preserved properties under transformations relevant to the problem domain - for example, translational symmetry in convolutional neural networks~\cite{lecun1998gradient} or permutation invariance in graph neural networks (GNNs)~\cite{scarselli2008graph}. Scale separation encapsulates the principle that interactions in many physical and informational systems are primarily local, allowing for hierarchical processing of information at different scales. GNNs for instance use the message-passing mechanism~\cite{gilmer2017neural} to aggregate information from local neighborhoods, with each layer potentially expanding the receptive field to capture multi-scale interactions. Transformers~\cite{vaswani2017attention}, which have revolutionized natural language processing~\cite{brown2020language} and computer vision~\cite{dosovitskiy2020image} since their introduction, achieving state-of-the-art performance across diverse domains from protein structure prediction to autonomous driving, can also be seen through the lens of geometric deep learning. Transformers are naturally permutation invariant and operate by implementing full pairwise attention between all nodes. This formulation allows Transformers to capture long-range dependencies directly, though at the cost of quadratic computational complexity in the input length. Recent architectural innovations like sparse attention patterns~\cite{child2019generating} and linear attention mechanisms~\cite{katharopoulos2020transformers} attempt to balance this global interaction capability with computational efficiency. Driven by the remarkable success of Transformers across diverse domains, researchers have actively explored ways to incorporate attention into graph learning frameworks. While earlier attempts aimed at applying attention at a local edge level~\cite{velivckovic2017graph}, recent work has focused on developing true Graph Transformer architectures~\cite{dwivedi2020generalization} that can capture global interactions while respecting graph structure. These approaches face two key challenges: (1) incorporating structural information in the absence of a natural sequential ordering, and (2) managing the computational complexity of all-pair interactions in large graphs. Various solutions have emerged to address these challenges. For structural encoding, approaches range from spectral methods using the graph Laplacian eigendecomposition~\cite{kreuzer2021rethinking} to spatial techniques based on random walk statistics~\cite{yeh2023random}. Another practical approach is to leverage message-passing neural networks (MPNNs) to first aggregate local topological information before the global attention mechanism, effectively combining local and global processing~\cite{rampavsek2022recipe}. In terms of computational scalability, researchers have proposed various structure-aware sparse attention patterns~\cite{shirzad2023exphormer} or use linear attention variants. In this work, we combine MPNNs for local structural feature aggregation with linear attention for efficient global information exchange and long-range dependency modeling. 

\paragraph{Data-driven inverse physics and flow reconstruction}
Many high-dimensional physical systems exhibit an inherent low-dimensional structure, where the underlying dynamics evolve on a manifold of much lower dimensionality than the original state space~\cite{jain2022compute}. This property has driven the development of inverse problem solving techniques that exploit the intrinsic low-dimensionality to reconstruct full system states from limited measurements. Key methodologies in this area include proper orthogonal decomposition (POD), which identifies orthogonal basis functions that maximize the captured variance in the data~\cite{berkooz1993proper}, dynamic mode decomposition (DMD), which extracts spatiotemporal coherent structures through spectral analysis of the system's evolution operator~\cite{schmid2010dynamic} and proper generalized decomposition (PGD), which employs a generalized spectral expansion approach to construct separated variable representations of the solution~\cite{chinesta2013proper}. These classical approaches have been successfully used to reconstruct flow fields and predict system evolution from partial observations~\cite{podvin2006reconstruction}, however, they typically assume linear relationships between system states, limiting their effectiveness for strongly nonlinear dynamics where modern machine learning methods can better capture complex patterns. Recent work~\cite{erichson2020shallow} has demonstrated the efficacy of a shallow neural network in addressing these limitations, proposing a minimal architecture of only 2-3 layers that directly learns the mapping between sensor measurements and flow fields. This approach achieves superior reconstruction accuracy compared to traditional modal decomposition methods while requiring significantly fewer sensors, highlighting the potential of even simple deep learning architectures to capture complex nonlinear relationships in fluid systems. Yet, this method remains tied to the specific spatial discretizations used during training, a limitation that recent approaches aim to overcome by learning more general, discretization-independent representations of the underlying physical fields.



\paragraph{Discretization-independent deep learning for PDEs}
While deep learning methods are commonly used nowadays to tackle scientific problems characterized by Partial Differential Equations (PDEs), the majority of the currently deployed methods are still heavily dependent on grid-based discretization schemes. Indeed, some of the most prevalent scientific deep learning approaches are convolutional~\cite{thuerey2020deep, wandel2020learning} or fully connected neural networks~\cite{ling2016reynolds}, which require training on fixed sized grids. Such schemes usually suffer from poor generalization and aliasing artifacts when applied to input structures which differ too greatly from the training setup, severely limiting their usefulness for applications such as flow reconstruction for arbitrary geometries. While recent work has attempted to address these challenges through approaches such as spline kernels~\cite{wandel2022spline}, neural operators (NOs)~\cite{lu2019deeponet,kovachki2023neural} have emerged as a particularly promising framework, offering discretization-independent solutions to PDE-based problems. The growing popularity of NOs can be attributed to their ability to map between the input space and an infinite dimensional function space, given a finite training set of collected input-output observations of the problem and independently of the discretization. The Fourier Neural Operator (FNO) in particular has demonstrated strong performance in a number of PDE tasks~\cite{li2020fourier}. This method works by projecting the input function onto the Fourier space, applying learned linear transformations in the frequency domain, and combining this with a local convolution operation in physical space to capture both global and local dependencies in the solution field. Motivated by this, we propose a scheme that combines elements from both graph-based~\cite{li2020neural} and Transformer-based~\cite{cao2021choose} Neural Operators (NOs), sharing conceptual similarities with the Fourier Neural Operator's global-local processing strategy while operating directly on graph structures.

\paragraph{Flow reconstruction with geometric deep learning}
Given the advantages of geometric deep learning for learning complex functions on irregular domains, there have been multiple attempts at reconstructing flows using such methods in the literature. Graph Convolutional Networks (GCN)~\cite{welling2016semi}, for instance, were used in \cite{chen2021graph} to reconstruct flows around random Bezier-curve shapes, albeit at a very low Reynolds number of $\mathrm{Re}=10$. In another GCN-based approach~\cite{he2022flow}, the flow behind a cylinder was reconstructed using inputs from randomly placed sensor points in the fluid domain, although also limited to low Reynolds numbers. As discussed in \autoref{sec:challenges}, our work builds on a previous attempt to reconstruct flows around airfoils using very deep reversible GNNs~\cite{duthe2023graph}. This previous work also introduced the use of Feature Propagation (FP)~\cite{rossi2021unreasonable} to initialize the features of the unknown nodes in the graph (which we also use) and compared multiple message-passing operators: a GCN-based approach, one based on Graph ATtention (GAT)~\cite{velivckovic2017graph} and another using the Graph Isomorphism Network (GIN)~\cite{xu2018powerful}. The effectiveness of the FP and GAT combination has been further validated by its successful adaptation to time-varying flow reconstruction in internal combustion engines~\cite{danciu2024flow}. Finally, GNNs have also been applied to the reconstruction of hypersonic flows close to aircraft~\cite{li2024novel}, where graph pruning strategies and optimized edge weights were used in order to reduce computational complexity.

\section{Generating a robust training dataset}
\label{sec:dataset}
This section describes in detail the generation process of our training dataset, which is based on use of diverse airfoil geometries. In total we obtain 2907 converged simulations that are turned into graphs, with 2469 used for training, 239 for validation and 199 for testing. The full graph training dataset can be accessed here: \url{https://doi.org/10.5281/zenodo.14629208}. The code used for our airfoil simulation pipeline is also made available and can be accessed at \url{https://github.com/gduthe/airfoil_sim_pipeline}.

\subsection{Simulation pipeline}
\begin{figure}
  \centering
  \includegraphics[width=0.9\linewidth]{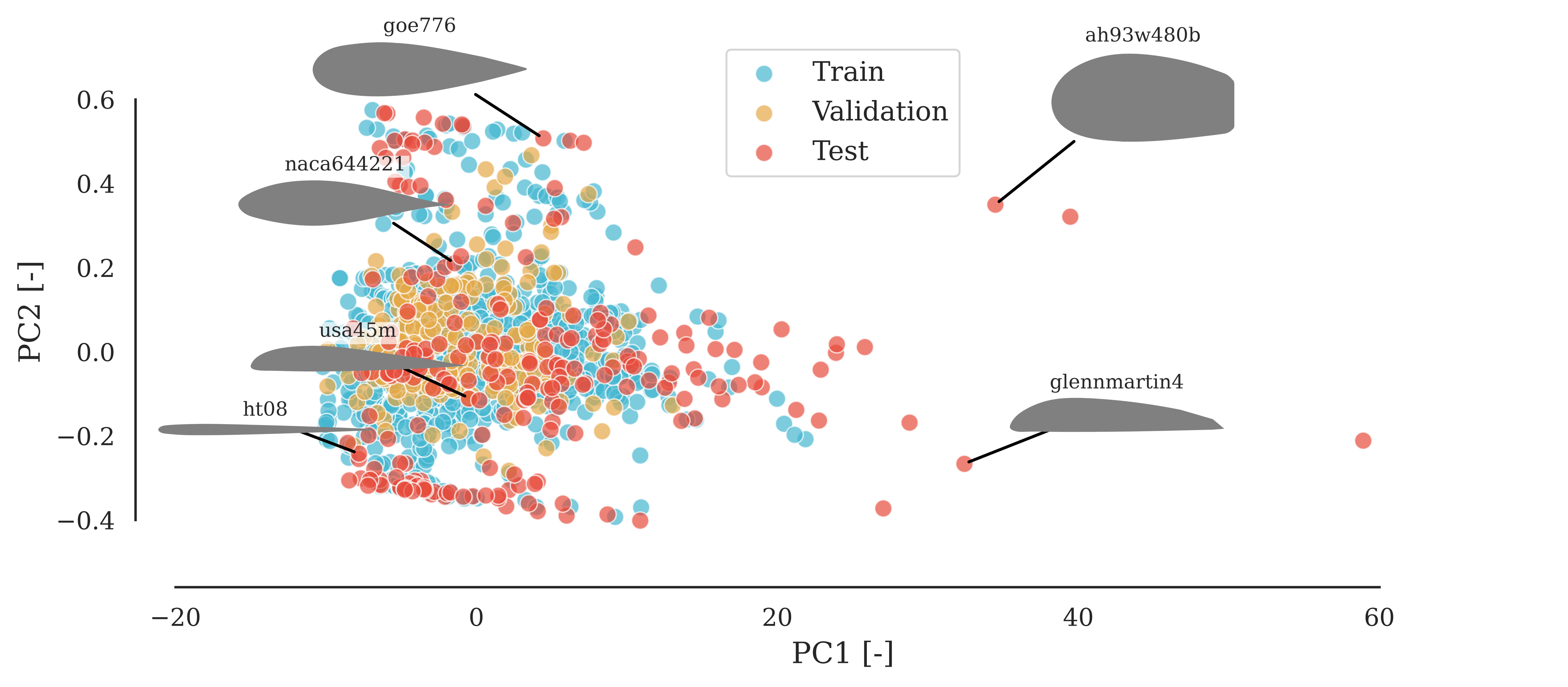}
  \caption{Distribution of airfoil geometries visualized through their first two principal components. The training set (blue) and validation set (orange) span the primary shape variations, while the test set (red) holds an increased share of outlier shapes. We plot some examples of the test set to showcase some of the outlier shapes.}
  \label{fig:airfoils}
\end{figure}

\paragraph{Airfoil geometry selection}
We obtain our airfoil geometries from the UIUC airfoil database \cite{selig1996uiuc}, applying several preprocessing steps to ensure data quality and representativeness. An initial cleaning phase removes airfoils with self-intersecting geometries and normalizes all shapes to a unit chord length. To better capture the critical aerodynamic features near the leading edge, we enhance the geometric resolution of the front half profiles through cubic spline interpolation. To prevent data leakage between evaluation sets, we partition the processed database into training, validation, and test sets \textit{prior to} conducting the aerodynamic simulations. We use a stratified approach to partition the cleaned database into training, validation, and testing sets: while the training and validation sets span the primary shape variations, we specifically reserve geometrically challenging airfoils for testing. These challenging cases include outlier geometries characterized by extreme camber, thickness ratios, or other distinctive features. \autoref{fig:airfoils} displays the distribution of airfoils across these three sets through a Principal Component Analysis (PCA) of the key geometric parameters.

\paragraph{Meshing} For mesh generation, we employ Gmsh \cite{geuzaine2020three} to create unstructured O-grid meshes around each airfoil geometry. We use a mesh refinement strategy that incorporates two key controls: a near-wall sizing field to maintain appropriate y+ values for CFD wall-function implementation, and a global sizing parameter to regulate far-field resolution while constraining the total cell count below 150,000 elements. Our choice of unstructured triangular elements offers two principal advantages: robust mesh generation across diverse geometries and the potential for flexible local refinement through adaptive point insertion and retriangulation. This meshing approach provides an effective balance between computational efficiency and solution accuracy while maintaining generalizability.

\paragraph{Boundary condition generation}
We implement a systematic sampling strategy for generating physically consistent boundary conditions for subsonic aerodynamics. This approach is inspired by domain randomization~\cite{tobin2017domain}, where diverse simulated training data enable models to generalize to real-world scenarios. The flow conditions span uniform distributions of inflow velocities (1-100 m/s) and angles of attack ($-20^\circ$ to $20^\circ$  for the training set and $-25^\circ$ to $25^\circ$ for the test and validation sets), maintaining incompressible flow regimes ($\text{Ma} < 0.3$). Turbulence characteristics are parameterized through intensity (uniform: 0-20\%) and length scales (uniform: 0.01-0.5 m), reflecting typical atmospheric conditions. Ambient air properties are sampled from a normal distribution (mean of $15^\circ$C, variance of $4.5^\circ$C), from which the dynamic viscosity is computed. The resulting Reynolds number distribution spans from $2\cdot10^5$ to $6.5\cdot10^6$, with a mean of $3\cdot10^6$, ensuring predominantly turbulent flow conditions. Our overall sampling strategy is based on Sobol sequences, which ensures coverage of the high-dimensional parameter space. 

\paragraph{CFD simulations} Each airfoil mesh is simulated twice, each time with a different inflow configuration drawn from the pool of boundary conditions generated as described above. Our simulation pipeline uses the OpenFOAM CFD software package~\cite{jasak2007openfoam} to solve the steady 2-D Reynolds-Averaged Navier–Stokes (RANS) equations, which can be written in Einstein notation as:
\begin{equation}
 \rho\left( \frac{\partial \overline{U_i}}{\partial t} + \frac{\partial}{\partial x_j}(\overline{U_i}\overline{U_j}) \right) = \frac{\partial p}{\partial x_i}  + \frac{\partial}{\partial x_j}\left(\mu \frac{\partial \overline{U_i}}{\partial x_j} \right)+ \frac{\partial}{\partial x_j}(-\rho \overline{u'_i u'_j})
\end{equation}
\noindent where $\overline{U_i}$ represents the mean velocity component in direction $i$, $\rho$ is the fluid density, $p$ denotes the pressure, $\mu$ is the dynamic viscosity, and $-\rho \overline{u'_i u'_j}$ represents the Reynolds stress tensor arising from the averaging of turbulent fluctuations. The Reynolds stress tensor is modeled using the Boussinesq approximation:
\begin{equation}
    -\rho \overline{u'_i u'_j} = \mu_t \Big( \frac{\partial U_i}{\partial x_j} + \frac{\partial U_j}{\partial x_i} \Big) - \frac{2}{3} \rho k \delta_{ij}
\end{equation}{}
where $\mu_t$ is the turbulent eddy viscosity, $k$ is the turbulent kinetic energy, and $\delta_{ij}$ is the Kronecker delta. We can obtain this term by using the $k$-$\omega$ SST turbulence closure model~\cite{menter2001elements}, which solves for the turbulent kinetic energy $k$ and the specific dissipation rate $\omega$. Standard OpenFOAM wall functions are employed for near-wall treatment. Finally, solution convergence is enforced through residual monitoring, requiring all solved quantities (pressure, velocity components, $k$, and $\omega$) to achieve normalized residuals below $5\times10^{-5}$. Simulations that fail to meet these convergence criteria are excluded from the dataset.

\begin{figure}[ht]
     \begin{subfigure}[t]{0.49\textwidth}
         \centering
         \includegraphics[width=\linewidth]{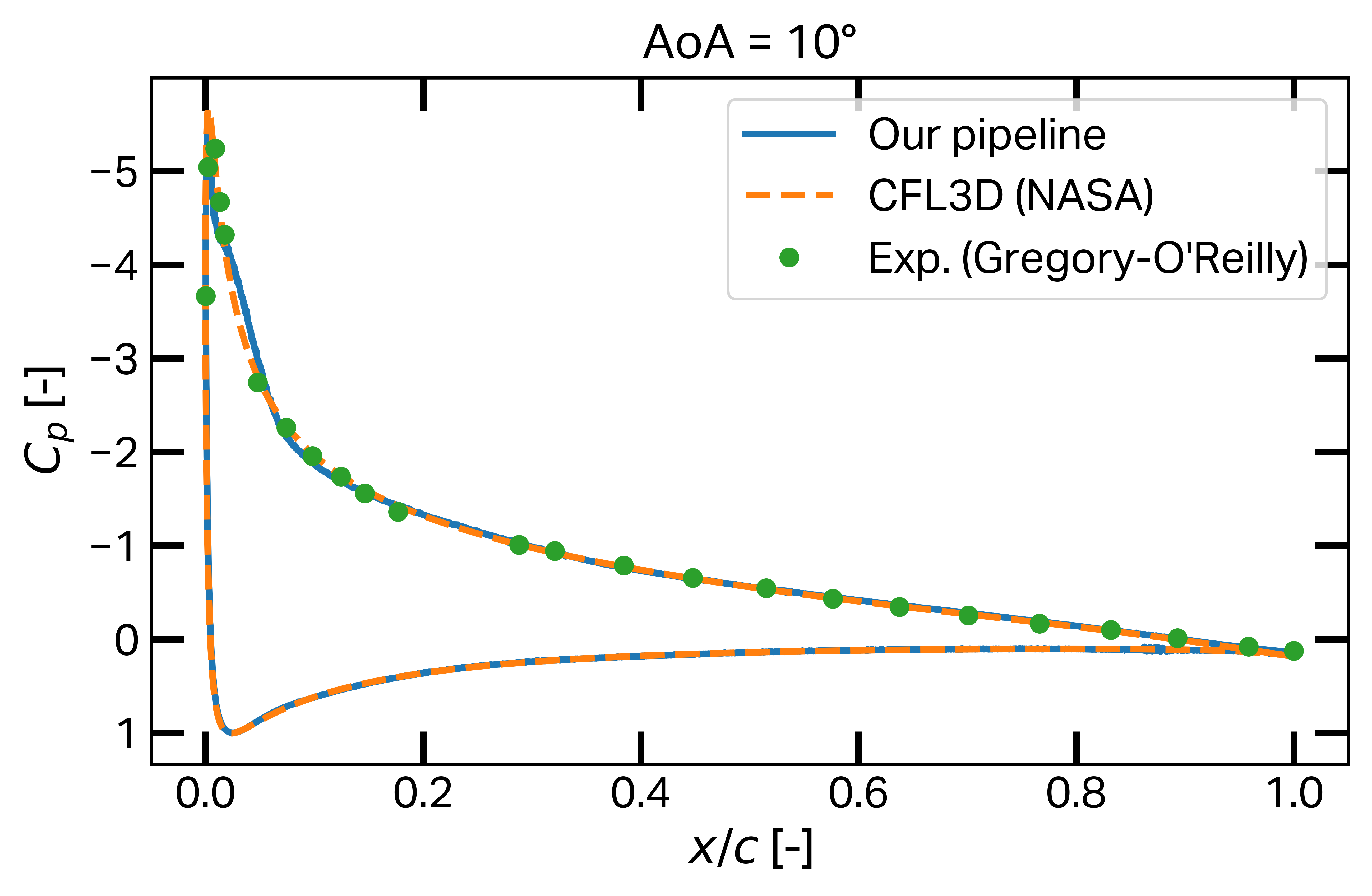}
         \caption{}
         \label{fig:cp_distrib10}
     \end{subfigure}
     \begin{subfigure}[t]{0.49\textwidth}
         \centering
         \includegraphics[width=\linewidth]{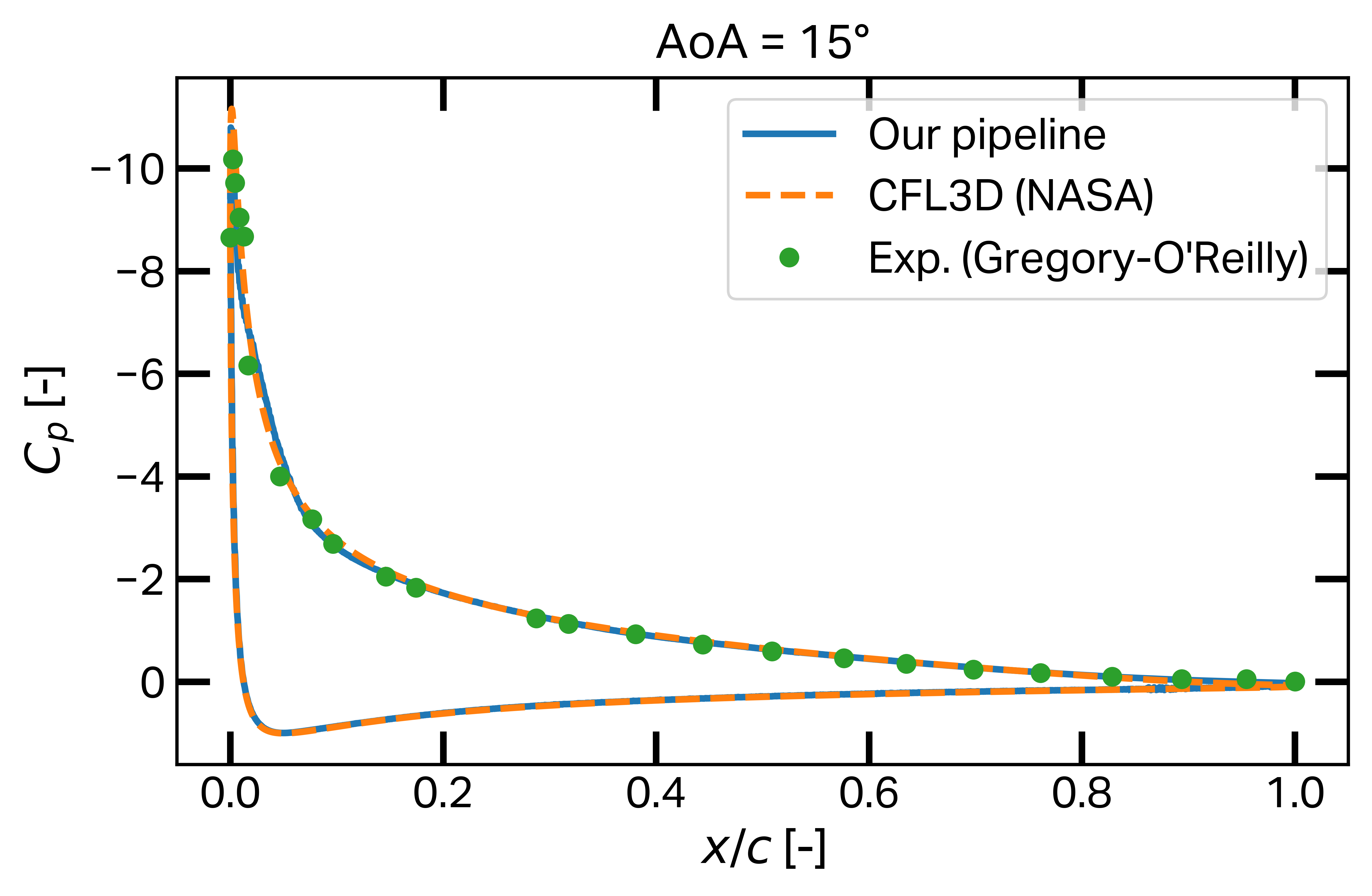}
         \caption{}
         \label{fig:cp_distrib15}
     \end{subfigure}
      \caption{Validation of the CFD pipeline against experimental measurements \citep{gregory1970low} and numerical simulations \citep{krist1998cfl3d}. The pressure coefficient distribution is shown for a NACA 0012 airfoil at $Re=6\cdot10^6$ for angles of attack of (a) 10$^\circ$ and (b) 15$^\circ$.}
  \label{fig:cp_distrib}
\end{figure}

To briefly validate our CFD pipeline, we use it to simulate a NACA 0012 airfoil and compare the results against established results~\citep{krist1998cfl3d, gregory1970low}. Figure~\ref{fig:cp_distrib} presents the pressure coefficient distributions at two distinct angles of attack, demonstrating good agreement between our simulations and both experimental and computational benchmarks from previous studies.

\subsection{Processing the simulations}
\begin{figure}[hb]
  \centering
  \includegraphics[width=1\linewidth]{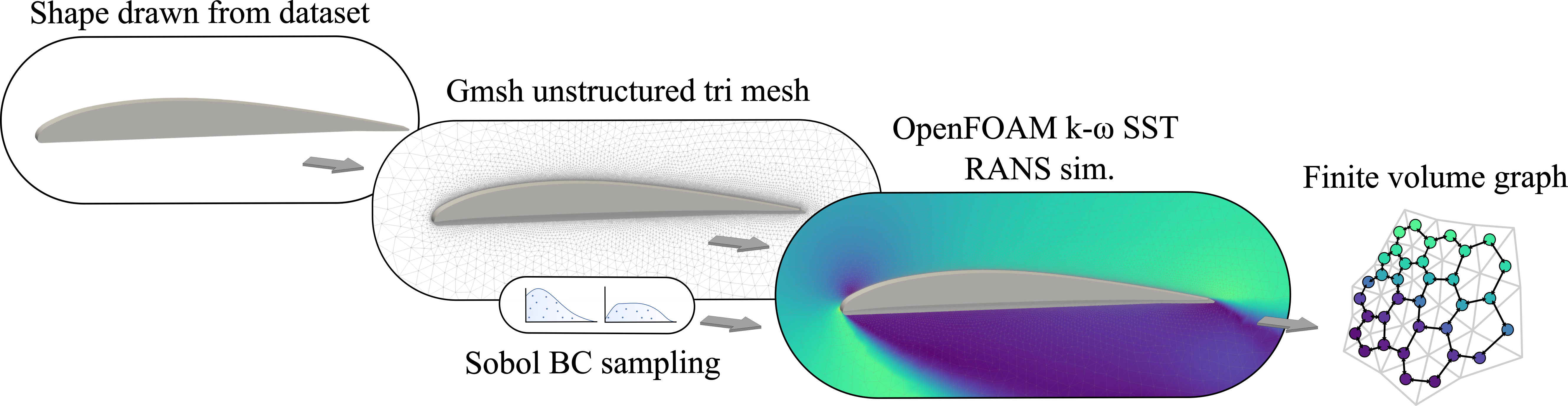}
  \caption{Illustration of the simulation pipeline. Airfoil shapes are selected at random from subsets of the UIC database, then meshed and simulated in OpenFOAM. We use quasi-random Sobol sequences to draw the boundary conditions such that the entire parameter range is filled. We use a finite-volume scheme to convert the simulations into the input graphs.}
  \label{fig:sim_pipeline}
\end{figure}

\paragraph{Mesh to graph translation}
We use a finite-volume inspired approach to convert our CFD meshes into graphs. In our implementation, mesh cells serve as nodes, with bidirectional edges connecting adjacent cells. This cell-centric representation also enables the extraction of physics-relevant edge features, specifically the boundary length $l_b$ between adjacent cells. This edge feature is useful, as it incorporates inherent cell sizing information. Cell quantities are converted into node features, comprising four components: pressure $p$, velocity components ($u_x$, $u_y$), and a one-hot encoded node category $\textbf{t}$ (fluid, or wall). To optimize computational efficiency, we restrict the graph representation to cells within one chord length of the airfoil, rather than processing the entire CFD domain (which extends to 100 chord lengths). The airfoil surface is represented by nodes positioned along its boundary, with bidirectional edges connecting adjacent surface nodes. These additional connections prevent the formation of tree-like structures that could impede learning performance~\cite{topping2021understanding}. The resulting graphs in our dataset contain on average approximately 55k nodes and 85k edges. \autoref{fig:sim_pipeline} illustrates the full simulation pipeline and the mesh-to-graph conversion scheme. 

\begin{figure}[hb]
  \centering
  \includegraphics[width=0.58\linewidth]{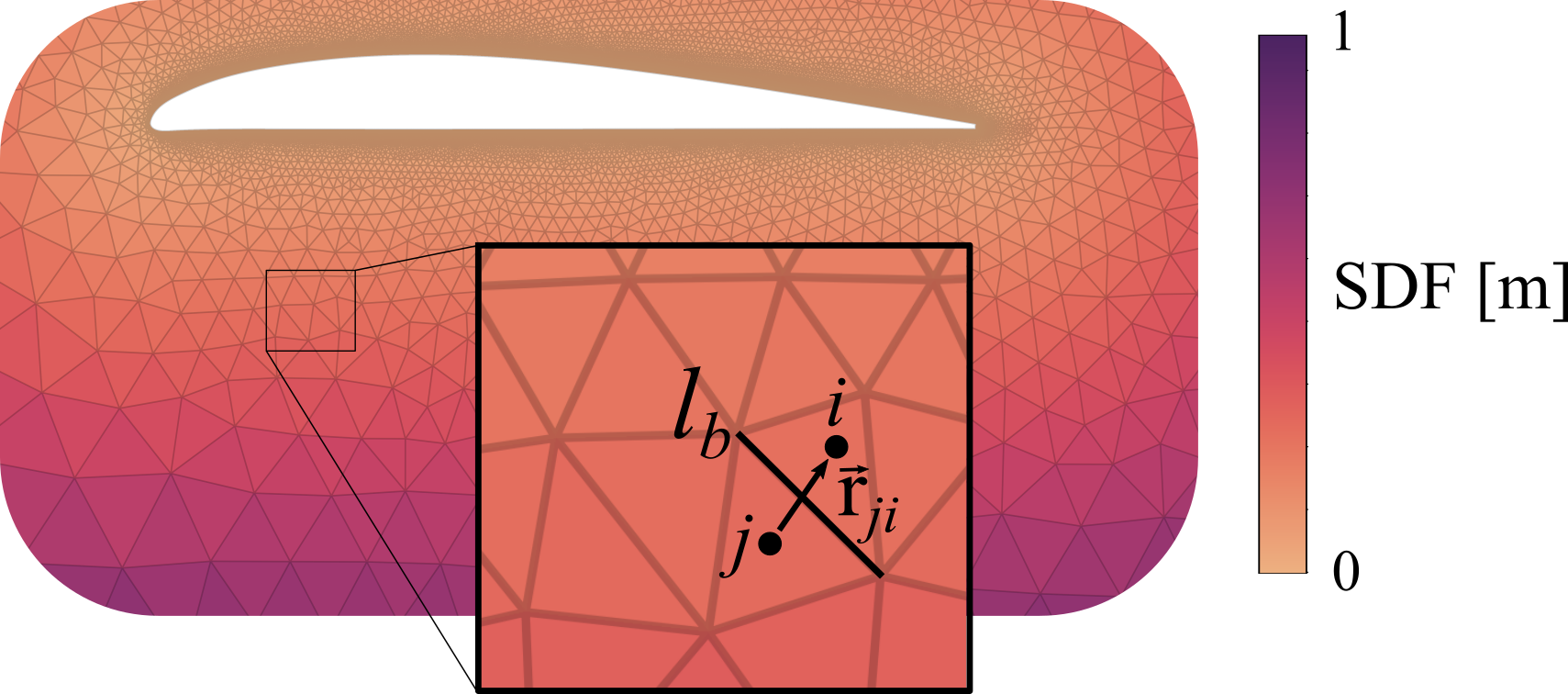}
  \caption{Geometrical features used as inputs to the model. Overall the signed distance field is plotted, with a zoom on the boundary length and relative edge direction between two nodes.}
  \label{fig:features}
\end{figure}

\paragraph{Input and output features}
For the node features, we only have access to the pressure distribution at the surface of the airfoil, while it is set to 'NaN' values at the fluid nodes. As an additional feature, we compute the signed distance field (SDF) for each node. The SDF is a continuous scalar field that encodes the minimum distance to the airfoil surface, an example of which is displayed in \autoref{fig:features}. In the flow reconstruction context, SDFs can provide an important geometric prior that helps the network understand spatial relationships. In our processing pipeline, we implement an efficient vectorized algorithm based on KD-trees for nearest neighbor search, enabling fast SDF computation for all fluid nodes even for large meshes. The type of each node is also known and is encoded as a one-hot vector, bringing the total size of the input node features to four. 

To account for mesh geometry, each edge of the graph is associated with some features. The following four edge characteristics are used as inputs: the x and y components of the relative edge direction vector $\vec{r}$, the edge length $l$, and the cell boundary length $l_b$, as previously described and shown in \autoref{fig:features}. We provide a summary of the properties of the input and output features, in \autoref{tab:in_out_features}.

\renewcommand{\arraystretch}{0.8}
\begin{table}[h]
\centering
\caption{Input and output features of the graph-based flow reconstruction problem.}
\begin{tabular}{@{}llc@{}}
\toprule
\textbf{Feature} & \textbf{Description} & \textbf{Dimension} \\ \midrule
\textbf{Input node features} & & \\
\quad $p$ & Surface pressure (airfoil nodes only, NaN elsewhere) & 1 \\
\quad $\textbf{t}$ & One-hot encoding (airfoil/fluid) & 2 \\
\quad $\sigma$ & Signed distance function & 1 \\ \rule{0pt}{3ex} 
\textbf{Input edge features} & & \\
\quad $\vec{r}$ & Relative edge vector ($r_x$, $r_y$) & 2 \\
\quad $l$ & Edge length & 1 \\
\quad $l_b$ & Cell boundary length & 1 \\ \rule{0pt}{3ex} 
\textbf{Output node features} & & \\
\quad $p$ & Pressure field & 1 \\
\quad $\vec{u}$ & 2D velocity vector ($u_x$, $u_y$) & 2 \\ \bottomrule
\end{tabular}
\label{tab:in_out_features}
\end{table}

\paragraph{Data normalization}
In our data preprocessing approach, we implement a normalization scheme for both input and target variables. The surface pressure measurements are normalized using their statistical moments - specifically the mean ($\mu_p$) and standard deviation ($\sigma_p$) computed from the measured pressure distribution along the airfoil surface. The same normalization is applied to the target pressure field. For velocity predictions, we normalize the target values by scaling them with respect to the estimated inflow velocity ($\hat{U}_{\infty}$), which can be computed using Bernoulli's principle. Given that all airfoils are simulated with a zero farfield static pressure, the farfield inflow velocity magnitude can be approximated as: 
\begin{equation}
    \hat{U}_{\infty} =  \sqrt{\frac{2 \cdot p_0}{\rho}}
\end{equation}
where $\rho$ is the density of air (constant throughout simulations) and $p_0$ is the total pressure measured at the stagnation point, which can be estimated by taking the maximum pressure at the airfoil nodes $ p_0 = \max(p_{_{\mathcal{V}_a}})$. While Bernoulli's principle is not valid for turbulent flows such as the ones we try to reconstruct, it serves as a good proxy for normalization purposes.
A key advantage of this normalization strategy lies in its exclusive reliance on surface measurements to compute the normalization parameters. This design choice potentially enables the model to reconstruct flows beyond the range of inflow conditions present in the training dataset, as all normalizing quantities are derived solely from observable surface data. This could be particularly promising for practical applications where direct measurement of freestream conditions may be unavailable.

\section{Hybrid Graph Transformers for inverse physics on meshes}
\label{sec:method}
\begin{figure}[h]
  \centering
  \includegraphics[width=1\linewidth]{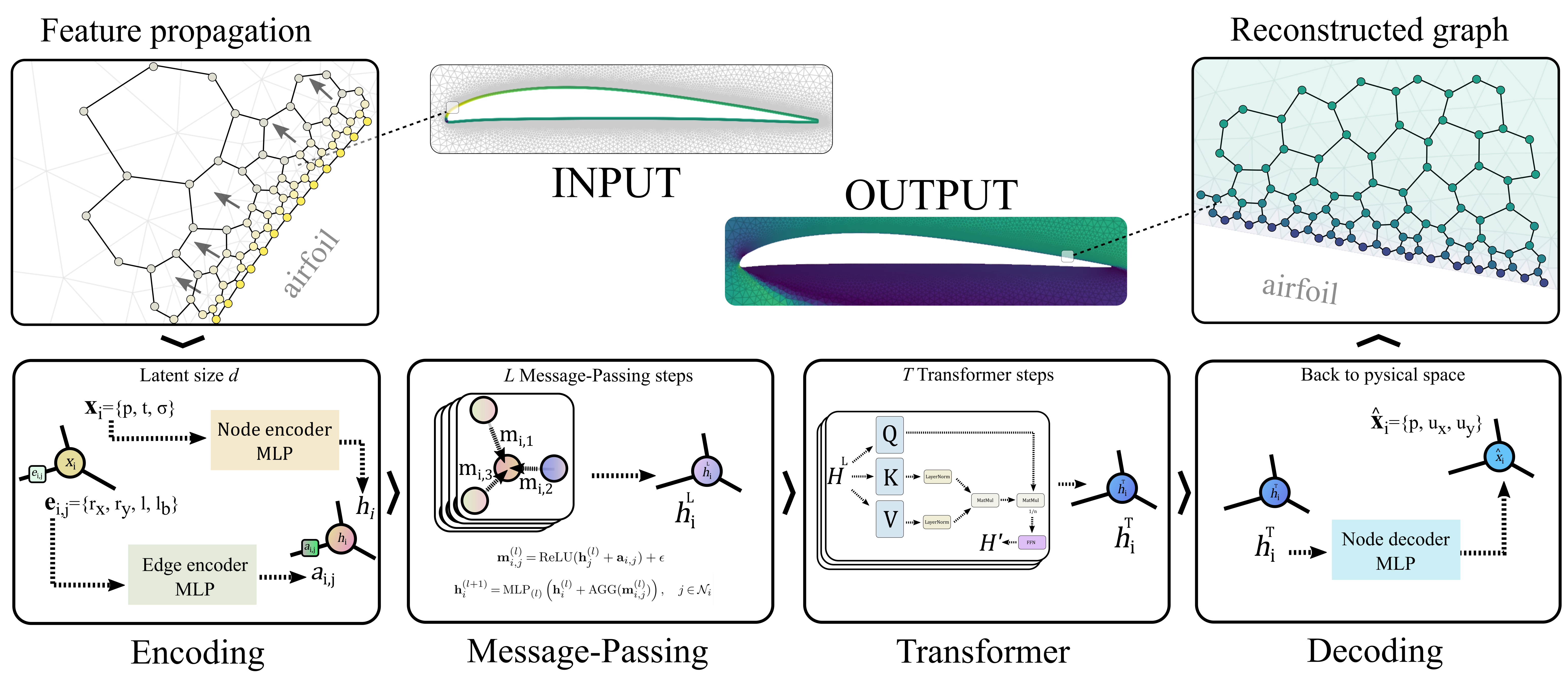}
  \caption{Architecture of the FRGT-S model. The pipeline consists of: (1) initial feature propagation on the graph, (2) node and edge encoding via MLPs, (3) iterative edge feature-aware message-passing layers for local information exchange, (4) Transformer layers for global information transmission, and (5) node-wise decoding to reconstruct the target flow field.}
  \label{fig:model_archi}
\end{figure}

In this section we describe the components which make up our Flow Reconstruction Graph Transformer (FRGT) model. The overall architecture is shown in \autoref{fig:model_archi}. The codebase for our model is made public and can be accessed at \url{https://github.com/gduthe/FRGT}.

\subsection{Overview of the architecture}
As a preliminary step, we use Feature Propagation~\citep{rossi2021unreasonable}, a matrix interpolation algorithm, to initialize the unknown node features. This step is an important part of our framework as it conditions an input graph into a plausible initial state, essentially radiating the surface node features outwards. We found 30 feature propagation iterations to be sufficient.

For the rest of our graph-based reconstruction framework, we adopt the now-popular Encode-Process-Decode graph-learning architecture~\cite{sanchez2018graph, pfaff2020learning}. The main idea behind this approach is to project the input features into a higher-dimensional latent space of size $d$ where the message-passing and attention operations can learn more expressive representations. This is particularly important in our setting, as the input features at each node typically consist of only a few physical quantities (pressure, velocity, node type and signed distance function), which limits the model's capacity to learn complex interactions in the original feature space. By first encoding these low-dimensional node features into a richer latent space, the subsequent processing steps can capture more nuanced relationships between nodes, before the decoder maps these learned representations back to the desired output quantities. This approach has been shown to effectively address the expressivity bottleneck inherent in working directly with low-dimensional physical variables in a number of problem settings~\cite{pfaff2020learning, hernandez2022thermodynamics}. 

The Encoding and Decoding layers consist of feed-forward neural networks that operate on each node/edge of the graph. For the encoding layer, two ReLU-activated multilayer perceptrons (MLPs) with three hidden layers each are used: an edge encoder that transforms the input edge features $\mathit{e}_{i, j}$ into encoded vectors $\mathbf{a}_{i, j}$ and a node encoder that processes the nodal input vectors $\mathbf{x}_i$ and global inputs $\mathbf{W}$ to generate the initial latent node features $\mathbf{h}_i$. This is expressed as:
\begin{align}
\mathbf{h}_{i} = \mathrm{MLP}_{ENC, node} (\mathbf{x}_i, \mathbf{W}), \quad \mathbf{a}_{i, j} = \mathrm{MLP}_{ENC, edge} (\mathbf{e}_{i, j})
\end{align}
Following the processing steps, a structurally similar decoding layer with a ReLU-activated MLP translates the high-dimensional transformed latent node features $\mathbf{h}_i'$ back into physically interpretable quantities:
\begin{equation}
    \hat{\mathbf{x}_{i}} = \mathrm{MLP}_{DEC, node} ( \mathbf{h}_i' )
\end{equation}


\subsection{Graph Transformer processor}
We introduce here the setup of the hybrid Graph Transformer processor. Similar to the approach outlined in \cite{rampavsek2022recipe}, we choose to use MPNNs to gather structural and geometrical information of the graph into informative node features before feeding them to a Transformer layer. This choice is motivated by several considerations: (1) MPNNs excel at efficiently processing large scale graphs, (2) geometrical edge attributes can easily be incorporated into message-passing allowing us to effectively encode geometric attributes such as relative distances, angles, and other spatial relationships between connected nodes, and (3) MPNNs circumvent the challenging problem of designing effective positional encodings for Graph Transformers, which remains an active area of research with multiple competing approaches.

\paragraph{Gathering local information via message-passing}
MPNNs operate by iteratively updating node representations through information exchange along edges, capturing both local structural patterns and geometric relationships. Each message-passing layer follows a dual-phase approach: first computing messages $\mathbf{m}_{i,j}$ between connected nodes, then aggregating these messages at each node through a pooling operation which is then fed through a function to update each node. In our work, we use the GENeralized Aggregation Networks (GEN) message-passing formulation~\citep{li2020deepergcn}, which extends the classical GCN to support edge features and more sophisticated message aggregation functions. This formulation has proven to be successful when applied to physics-based problems that depend on geometrical features~\cite{duthe2024flexible, de2024multivariate}. A single GEN update layer $l$ consists of a message computation phase incorporating edge attributes $\mathbf{a}_{i,j}$:
\begin{equation}
    \mathbf{m}_{i, j}^{(l)} = \mathrm{ReLU} (\mathbf{h}_{j}^{(l)} + \mathbf{a}_{i, j}) +\epsilon
\end{equation}
\noindent followed by an aggregation step that uses a learnable \texttt{softmax} function to weight the importance of different messages in the node update function:
\begin{align}
    \mathrm{AGG}(\mathbf{m}_{i, j}^{(l)})= \sum_{j \in \mathcal{N}_i} \frac{\exp(\beta \cdot \mathbf{m}_{i,j}^{(l)})}{\sum_{k \in \mathcal{N}_i} \exp(\beta \cdot \mathbf{m}_{i,k}^{(l)})} \cdot \mathbf{m}_{i,j}^{(l)} \\
    \mathbf{h}_i^{(l+1)} = \mathrm{MLP}_{(l)} \left( \mathbf{h}_i^{(l)} +
        \mathrm{AGG}(\mathbf{m}_{i, j}^{(l)})  \right), \quad j \in \mathcal{N}_i \quad 
\end{align}
\noindent This formulation allows the model to adaptively weigh the contributions of different neighbors based on their relevance, while maintaining numerical stability through the $\epsilon$ term (usually set to $10^{-7}$).

\paragraph{Efficient long-range transmission of information using a linear Transformer}
After the message-passing steps, the updated node features are fed into a Transfomer layer to capture long range interactions. Unlike the message-passing operations, which are constrained by the graph's connectivity, the self-attention mechanism in Transformers enables direct communication between all pairs of nodes, allowing the model to learn salient dependencies regardless of geometric distance or graph topology. To efficiently handle our large mesh-graphs, we opt for the Galerkin Transformer~\cite{cao2021choose}, a linear variant specifically optimized for PDE-related tasks. The key innovation in this approach lies in its normalization scheme and simplified attention mechanism, which can be expressed as:
\begin{equation}
z_i = \mathrm{ATTN} (h_i) = Q(K^T V)/n
\end{equation}
\noindent where $Q$, $K$ and $V$ are respectively the query, key and value vectors obtained via learned linear projections, and $n$ denotes the number of nodes in the graph. By replacing the traditional softmax normalization with a Galerkin projection-based layer normalization scheme, this Neural Operator (NO) is both accurate and computationally efficient, with empirical evaluations demonstrating strong performance across diverse PDE applications. In practice, we use $\eta$ attention heads in parallel, where each head independently processes the input latent node feature matrix $H$ into representations for each node of dimensionality $d_{\eta}$, which are then concatenated:
\begin{equation}
\mathrm{MultiHead}(H) = \mathrm{Concat}(head_1, ..., head_{\eta})W^O
\end{equation}
\noindent where $head_j = \mathrm{ATTN}(HW^Q_{\eta}, HW^K_{\eta}, H W^V_{\eta})$. $W^Q_{\eta}$, $W^K_{\eta}$, $W^V_{\eta}$ are learned weight matrices specific to head $\eta$ and $W^O$ is an output projection matrix. This multi-headed architecture enables the model to jointly attend to information from different representation subspaces at different positions, increasing its expressivity.

\begin{figure}[h]
  \centering
  \includegraphics[width=0.6\linewidth]{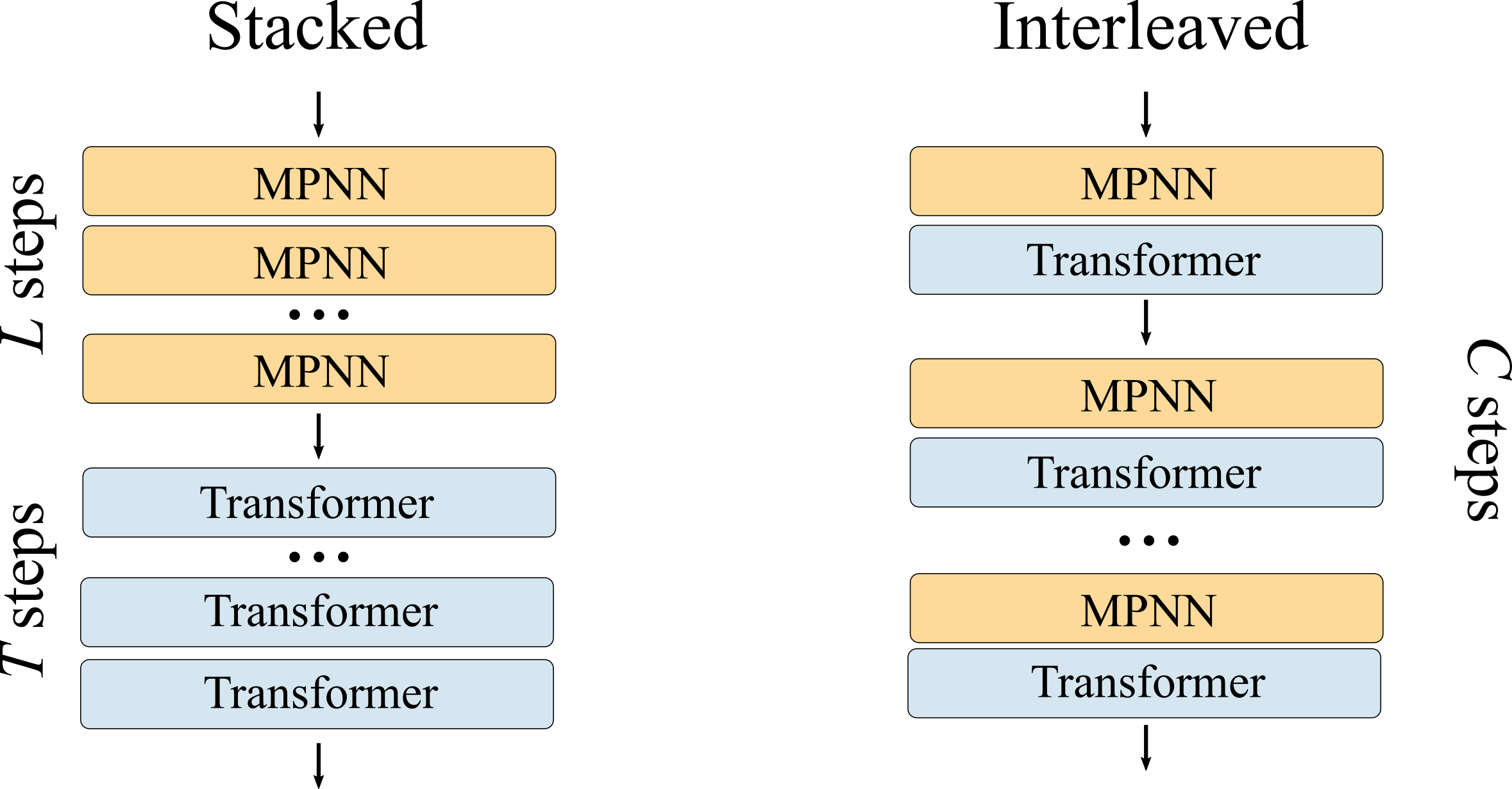}
  \caption{Illustration of the two proposed approaches for combining local and global layers.}
  \label{fig:layer_comp}
\end{figure}

\paragraph{Combining local and global layers}
The local MPNN and global Transformer layers could be combined using a number of different integration strategies. In this work, we consider first an approach that follows a sequential design where $L$ message-passing layers are stacked, followed by $T$ Transformer layers, allowing the model to first capture local patterns before applying global corrections. An alternative approach, inspired by~\cite{rampavsek2022recipe}, interleaves $C$ combined message-passing and Transformer layers, potentially enabling more complex hierarchical feature extraction through the simultaneous consideration of local and global information at each level. \autoref{fig:layer_comp} illustrates these two approaches. We investigate both strategies and explore how these different architectural choices impact the model's reconstruction ability.

\section{Results}
\label{sec:results}
\subsection{Architectural experiments}
We conduct an initial set of experiments to systematically evaluate our model's performance and analyze the impact of key architectural decisions. In these experiments, we consider the idealized scenario where pressure measurements are available at \textit{all} airfoil surface nodes. To ensure fair comparison across architectural variants, we maintain a consistent model capacity of approximately 1.4M trainable parameters. This constraint is chosen such that the model's memory footprint for the largest graph of the dataset during training does not exceed the 24GB of VRAM of the RTX 4090 GPU that we use. All models are implemented using \texttt{PyTorch} and \texttt{PyTorch-Geometric} and are trained to minimize a $L_2$ loss under identical conditions. Specifically, we train the models for 500 epochs using the AdamW optimizer~\cite{loshchilov2017decoupled} with a weight decay of $1\times$10$^{-4}$ and an initial learning rate of $5\times$10$^{-4}$. The learning rate is annealed during training through a cosine decay function~\citep{loshchilov2016sgdr}. 

\subsubsection{Interleaved vs stacked layers}
The combination of local (MPNN) and global (Transformer) processing layers can follow different architectural patterns. We investigate two integration strategies: a sequential approach where MPNN layers are followed by Transformer layers, and an interleaved design that alternates between them. We benchmark both approaches against a pure message-passing baseline, which uses the Reversible GAT architecture from~\cite{duthe2023graph}. \autoref{tab:comp_results} displays reconstruction error metrics gathered on unseen airfoils of the test dataset for these three models. 

\renewcommand{\arraystretch}{0.8}
\begin{table}[h]
\caption{Reconstruction metrics for the FRGT-Stacked, FRGT-Interleaved and Reversible GAT models, averaged on the test dataset over 3 different model training runs with different initializations.}
\centering
\begin{tabular}{@{}lccc@{}}
\toprule
\textbf{Metric}  & \begin{tabular}[c]{@{}c@{}}\textbf{FRGT-Stacked}\\ \scriptsize{$L=10$, $T=1$, $d=160$}\end{tabular} & \begin{tabular}[c]{@{}c@{}}\textbf{FRGT-Interleaved}\\ \scriptsize{$C=5$, $d=160$}\end{tabular}  & \begin{tabular}[c]{@{}c@{}}\textbf{Reversible GAT}\\ \scriptsize{$L=35$, $d=180$}\end{tabular} \\ \midrule
\textbf{Pressure} & & & \\
\quad Avg. RMSE [Pa] & 86.79 ± 4.5 & 89.44 ± 4.7 & 75.42 ± 16.9 \\
\quad Max Abs. Error [Pa] & 1382.25 ± 86.8 & 1428.13 ± 85.7 & 1170.67 ± 100.8 \\
\quad Test $R^2$ & 0.997 & 0.996 & 0.997 \\ \rule{0pt}{2.5ex} 
\textbf{x-Velocity} & & & \\
\quad Avg. RMSE [m/s] & 3.09 ± 0.06 & 2.92 ± 0.07 &  4.16 ± 0.61 \\
\quad Max Abs. Error [m/s] & 26.74 ± 0.91 & 29.68 ± 0.23 & 29.28 ± 2.08 \\
\quad Test $R^2$ & 0.982 & 0.980 & 0.973 \\ \rule{0pt}{2.5ex} 
\textbf{y-Velocity} & & & \\ 
\quad Avg. RMSE [m/s] & 1.30 ± 0.02 & 1.39 ± 0.02 & 2.04 ± 0.42 \\
\quad Max Abs .Error [m/s] & 18.03 ± 0.26 & 20.42 ± 0.15 &  26.27 ± 3.98 \\
\quad Test $R^2$ & 0.988 & 0.983 & 0.976 \\ \midrule
\textbf{Num. params [M]} & 1.39 & 1.39 & 1.45 \\ 
\textbf{Avg. GPU inference time [ms]} & 198 & 223 & 791 \\ 
\bottomrule
\end{tabular}
\label{tab:comp_results}
\end{table}

\begin{figure}[h!]
  \centering
  \includegraphics[width=1\linewidth]{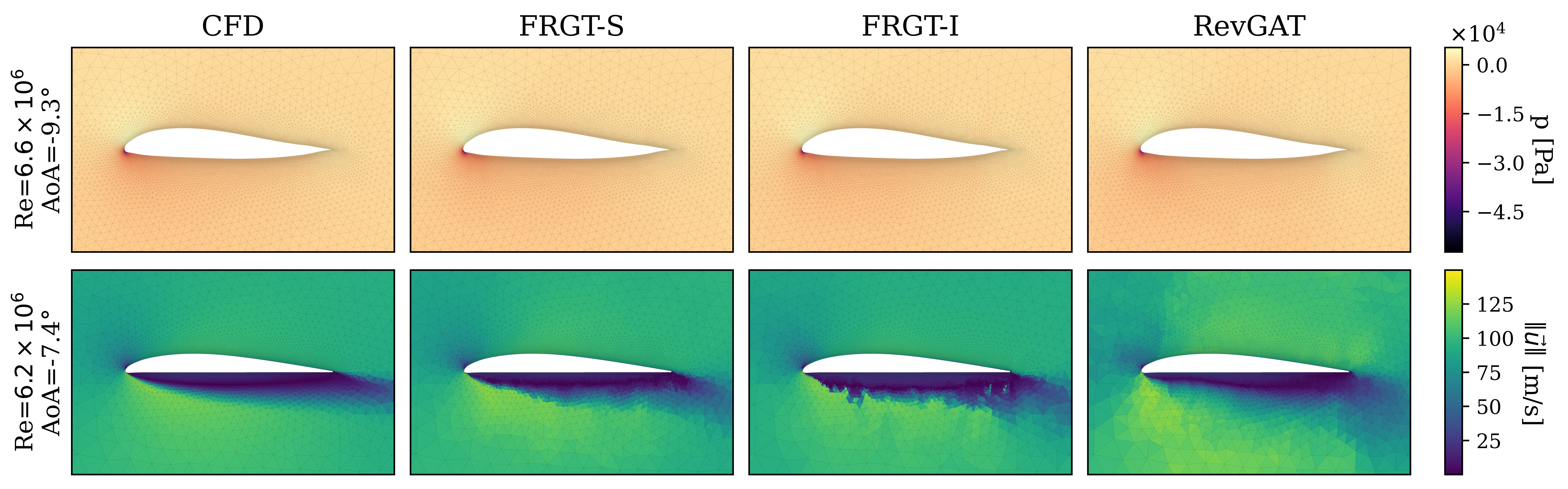}
  \caption{Comparing the reconstructed pressure and velocity magnitude of the FRGT-Stacked, FRGT-Intertwined and Reversible GAT~\cite{duthe2023graph} models against the ground truth CFD for two unseen airfoils in different inflow configurations. While all models can accurately reconstruct the surrounding pressure, the velocity recovery remains more challenging, especially for detached flows with strong shearing characteristics. The FRGT-S outperforms the other two models when this is the case. }
  \label{fig:model_comp1}
\end{figure}

The FRGT variants demonstrate good quantitative performance across measured metrics, with the interleaved and stacked approaches achieving comparable RMSE values for pressure reconstruction (86.79 ± 4.5 Pa and 89.44 ± 4.7 Pa respectively) and maintaining high $R^2$ values (>0.996). In velocity prediction, the interleaved architecture shows a slight advantage in x-velocity reconstruction (RMSE: 2.92 ± 0.07 m/s vs 3.09 ± 0.06 m/s), while the opposite is observed for the y-velocity component (1.30 ± 0.02 and 1.39 ± 0.02 m/s RMSE). Qualitative examination reveals that the stacked architecture, which processes local features through $L$ message-passing layers before applying $T$ transformer layers for global corrections, produces smoother flow field reconstructions with fewer artifacts, as shown in \autoref{fig:model_comp1}.

Both FRGT variants maintain similar computational characteristics with identical parameter counts (1.39M) and similar inference times ($\sim$200ms). The baseline Reversible GAT, while achieving competitive pressure reconstruction (75.42 ± 16.9 Pa RMSE), shows limitations in velocity field prediction, particularly for detached flows and shear layers. Notably, despite its marginally larger parameter count (1.45M), the Reversible GAT exhibits significantly higher computational overhead with an inference time of 791ms - approximately 4× slower than the FRGT variants. These limitations stem from its reliance on a reversible local message-passing scheme to propagate information from the surface outwards, whereas the Transformer-enhanced architectures can directly capture the long-range dependencies needed for accurate velocity field reconstruction in the more complex flow cases. The performance gap is most evident in regions requiring integration of information across larger spatial scales, demonstrating the benefits of combining both local and global processing mechanisms for flow field reconstruction and solving inverse physics problems using only boundary conditions. In light of these results, we select the FRGT-S model for subsequent experiments.

\subsubsection{Architecture design trade-offs for the FRGT-S model}
\begin{figure}[h!]
  \centering
  \includegraphics[width=0.95\linewidth]{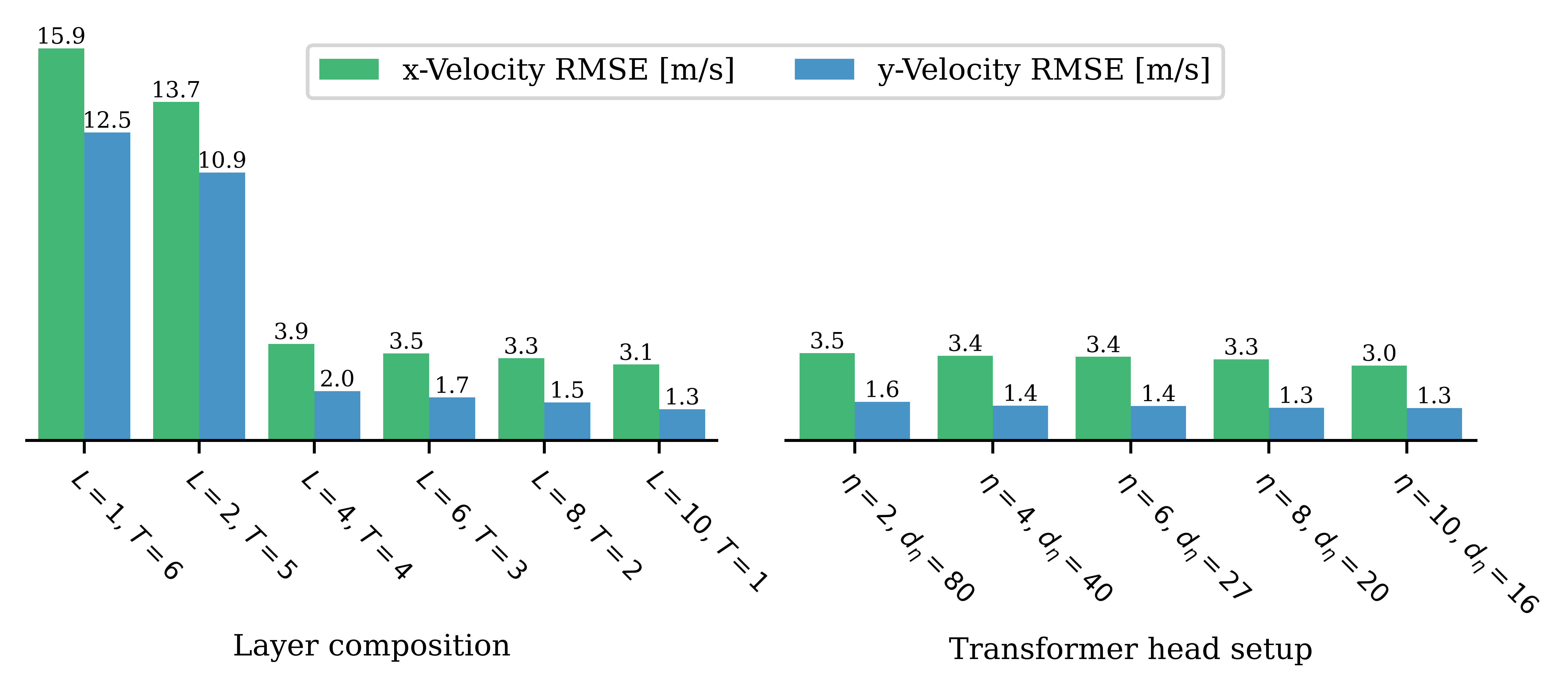}
  \caption{Study comparing model performance (x- and y-velocity RMSE) across message-passing vs. Transformer layer ratios ($L:T$) and attention head configurations ($\eta:d_{\eta}$). Lower RMSE indicates better performance, with layer composition showing stronger impact on accuracy than attention head setups.}
  \label{fig:model_opti}
\end{figure}

We then analyze key architectural trade-offs in our hybrid stacked Graph Transformer model through two studies, the results of which are shown in \autoref{fig:model_opti}. First, we examine the balance between local message-passing and global attention mechanisms while maintaining a fixed parameter budget (1.4M trainable parameters). 

The results show that increasing the number of message-passing layers $L$, while reducing the subsequent Transformer layers $T$, leads to significant improvements in reconstruction accuracy, with x-velocity RMSE decreasing from 15.9 m/s at $L=1, T=6$ to 3.1 m/s at $L=10, T=1$. This strong dependence on initial message-passing steps stems from our architectural choice of encoding geometric information only using a signed distance function (without additional positional encodings) and via the edge features, the latter of which the Transformer does not have access to. The second study investigates attention head configurations by varying the number of heads $\eta$ and head dimension $d_{\eta}$. The results indicate a preference for architectures with more, smaller heads over fewer, larger ones, though the impact on performance (x-velocity RMSE between 3.0-3.5 m/s) is less pronounced than layer composition choices. This aligns with established Transformer literature suggesting that multiple attention heads enable parallel processing of different input aspects. These findings emphasize the importance of sufficient local geometric processing through message-passing before applying global attention mechanisms for Graph Transformer approaches on meshes.

\subsection{Reconstruction of flows around unseen airfoils}
\begin{figure}[h!]
  \centering
  \includegraphics[width=1\linewidth]{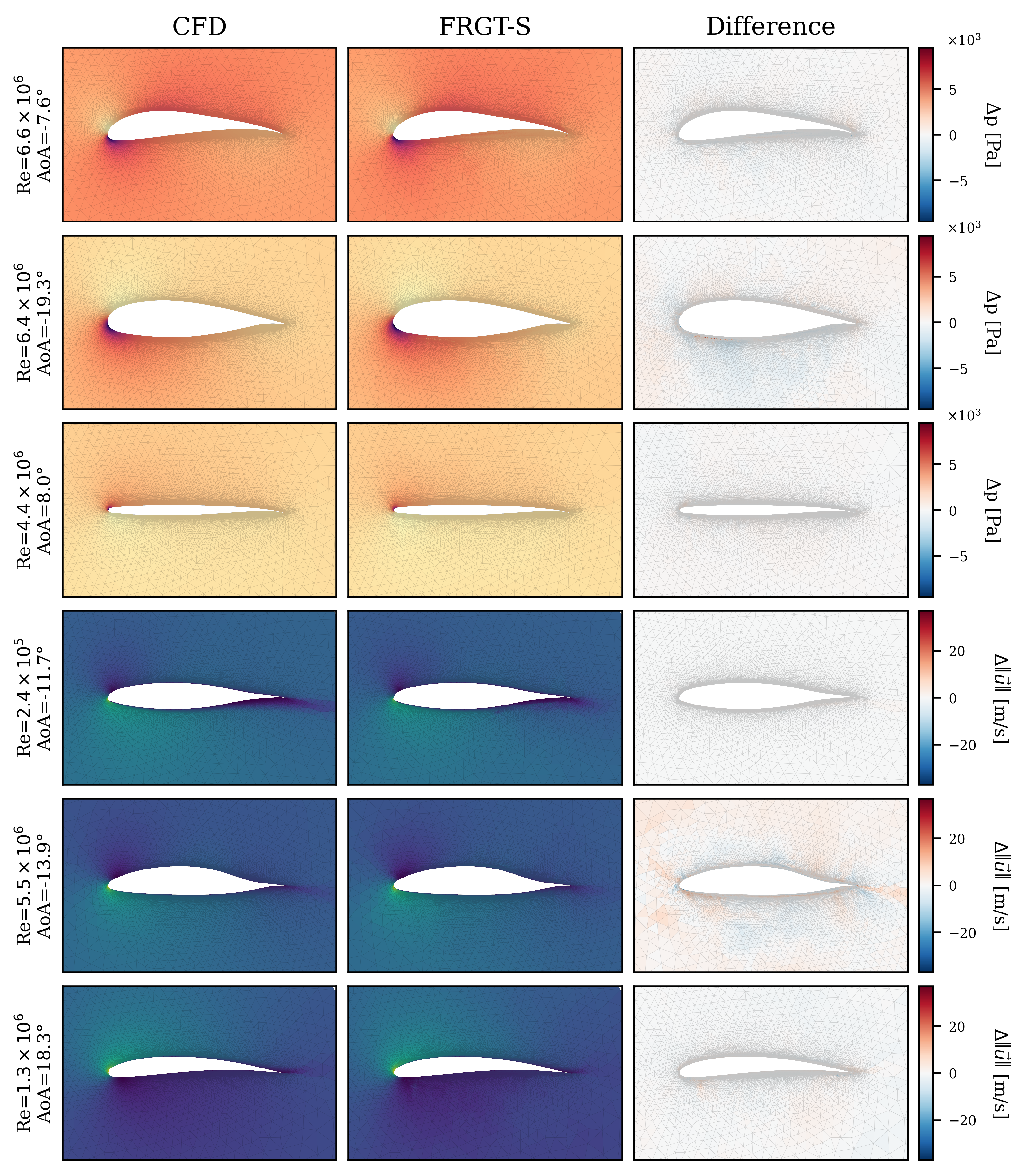}
  \caption{Comparison of flow field predictions between the CFD ground truth and the FRGT-S reconstructions for different airfoils across varying Reynolds numbers (Re) and angles of attack (AoA). The rightmost column shows prediction errors, demonstrating consistent accuracy across unseen airfoil geometries for inflow conditions within the training envelope.}
  \label{fig:quali_pred_easy}
\end{figure}

\autoref{fig:quali_pred_easy} presents qualitative results of our FRGT-S model's performance on unseen airfoils within the training distribution of the angle of attack ($-20^\circ$ to $20^\circ$). The visualization compares ground truth CFD solutions with reconstructed fields across six different test cases, spanning both pressure and velocity predictions at varying Reynolds numbers ($Re$) and angles of attack ($AoA$). The pressure field reconstructions (top three rows) demonstrate the model's ability to accurately capture the pressure distribution around different airfoil geometries. The difference plots reveal that errors remain small across the domain, with slight deviations primarily occurring in the near-wall region and wake area. The model successfully reproduces key flow features, such as the stagnation point at the leading edge and pressure gradients close to the airfoil surface. The velocity field predictions (bottom three rows) show equally promising results, with the model accurately reconstructing the flow patterns, including the velocity gradients in the boundary layer and wake regions. The difference plots indicate that the largest discrepancies appear in the wake region, where flow structures are more complex and sensitive to small perturbations. However, these errors remain within acceptable bounds ($\Delta u$ typically less than 5 m/s), particularly considering the challenging nature of velocity field reconstruction from surface measurements alone. Notably, the model maintains consistent performance across different Reynolds numbers and angles of attack, suggesting robust generalization to unseen airfoil geometries within the trained parameter space. 

\begin{figure}[h!]
  \centering
  \includegraphics[width=1\linewidth]{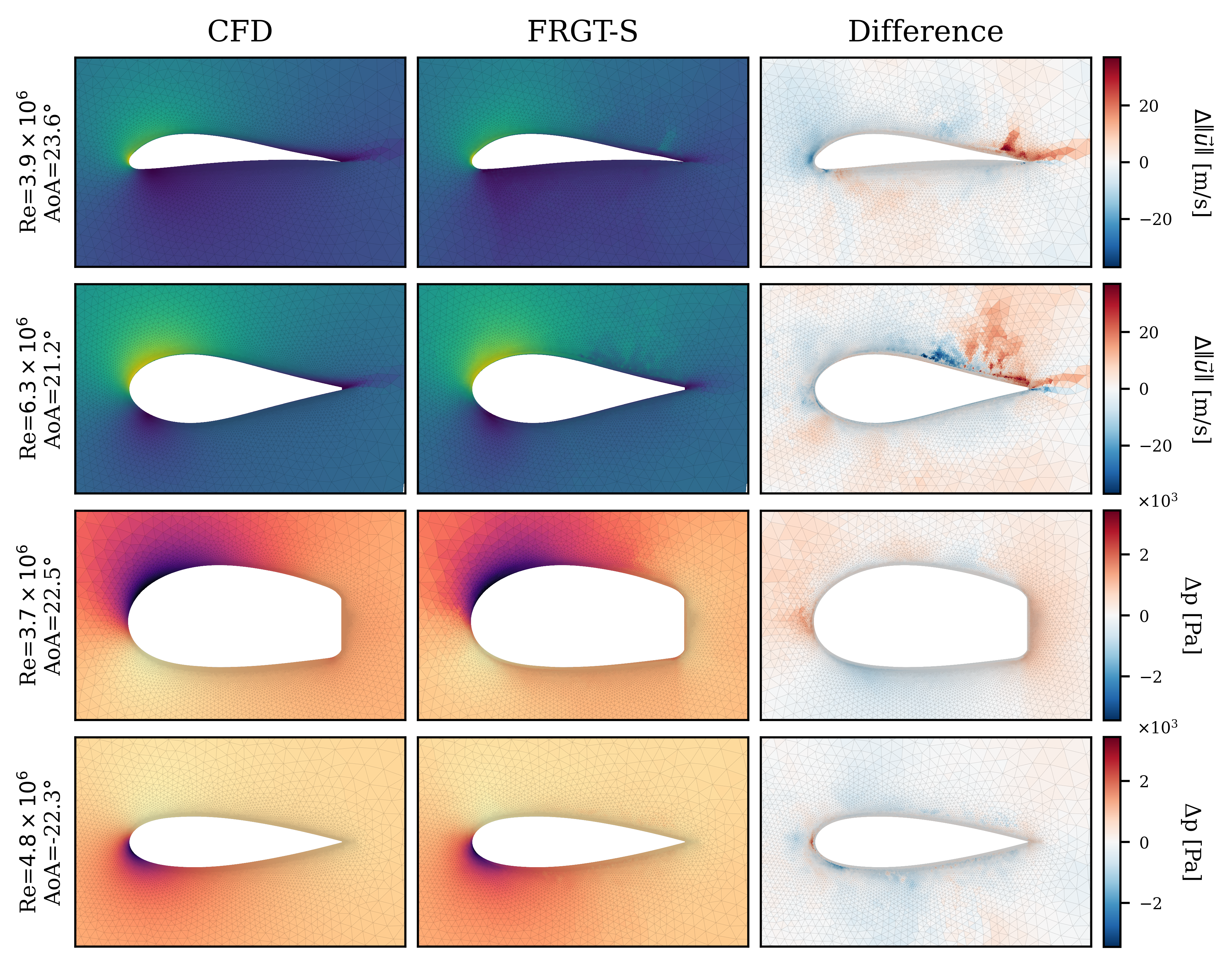}
  \caption{Evaluation of the FRGT-S model's extrapolative performance comparing  ground truth CFD solutions with predictions and showing prediction errors in the rightmost column. Test cases demonstrate generalization to unseen angles of attack (AoA < -20° and AoA > 20°) and unconventional geometries.}
  \label{fig:quali_pred_hard}
\end{figure}

\autoref{fig:quali_pred_hard} examines our model's performance under two challenging extrapolation scenarios: angles of attack outside the training range of $[-20^\circ, 20^\circ]$ and unconventional airfoil geometries. The test cases include thick, highly cambered profiles that differ significantly from the more conventional NACA-style airfoils that dominate the training dataset. At $AoA=23.6^\circ$ and $21.2^\circ$ (top two rows), despite the extreme angles and unusual airfoil shapes, the model captures the general flow structure, though with increasing errors in the wake prediction. The third row showcases a particularly challenging case - a very thick, blunt airfoil at $AoA=22.5^\circ$. Here, the difference plot reveals more substantial errors in pressure prediction, especially around the leading edge and in regions of expected flow separation. The bottom row ($AoA=-22.3^\circ$) demonstrates reasonable performance despite the large negative angle of attack. In general, errors are more pronounced compared to predictions on conventional airfoils within the training distribution.

\subsection{Partial airfoil coverage}
\begin{figure}[h!]
  \centering
  \includegraphics[trim={48cm 0 0 0},clip, width=1\linewidth]{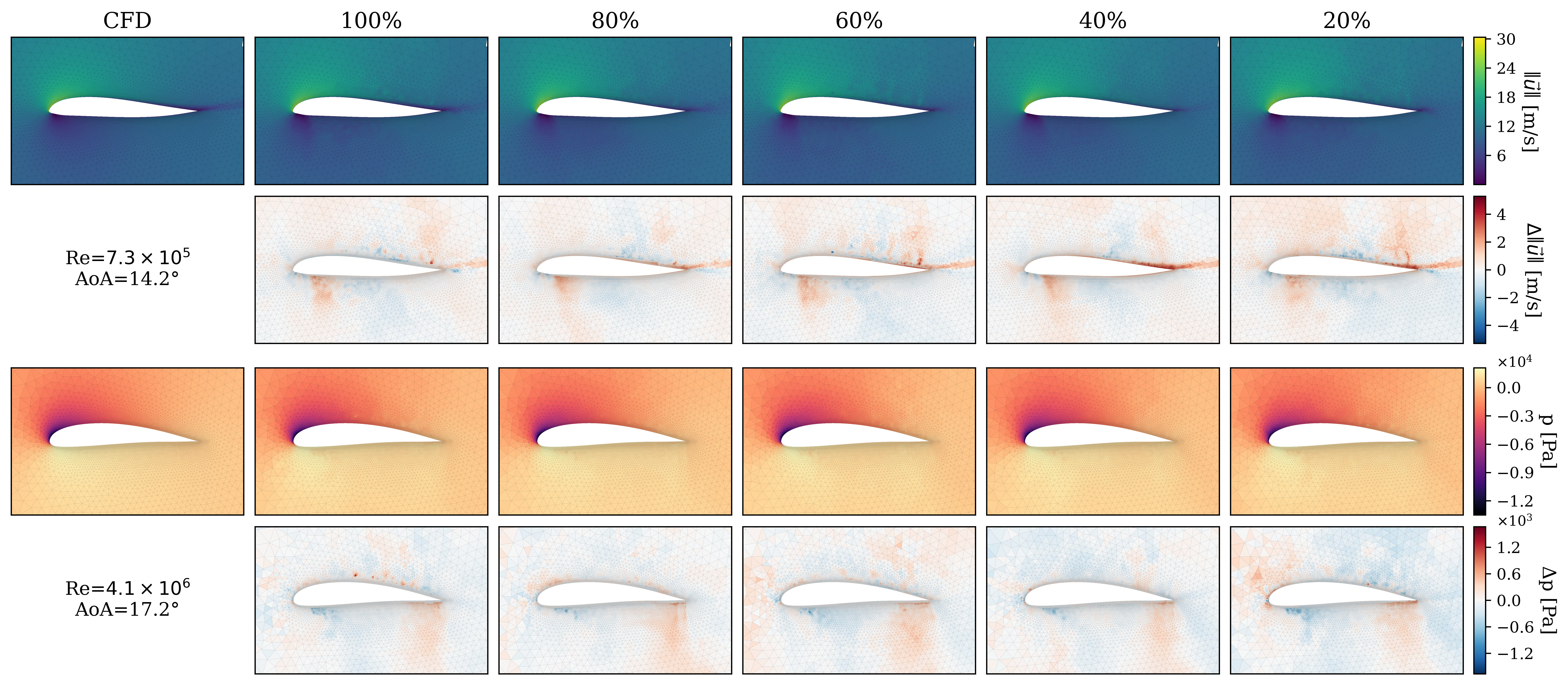}
  \caption{Comparison of flow field predictions (velocity magnitude and pressure) for two representative airfoils using FRGT-S models trained with varying degrees of surface sensor coverage (100\% to 20\%). Top rows show predicted fields, while bottom rows display deviations from CFD ground truth. Results demonstrate robust reconstruction capability even with significantly reduced sensor coverage, maintaining key flow features across different coverage scenarios.}
  \label{fig:partial_cov}
\end{figure}

Real-world aerodynamic sensing applications often face practical limitations in sensor coverage. For instance, the AeroSense MEMS-based pressure measurement system~\cite{barber2022development}, while offering robust in-field blade aerodynamics monitoring capabilities, is met with inherent coverage constraints due to electronics limitations. This is particularly relevant for large-scale wind turbine blades, where comprehensive surface coverage becomes increasingly challenging. To assess our method's viability under such realistic sensing conditions, we evaluate the performance of the FRGT-S model across varying degrees of sensor coverage. Our experimental setup simulates different sensor configurations by restricting pressure measurements to 20\%, 40\%, 60\%, and 80\% of the chord length along the front half of the airfoil. We train identical models for each partial coverage scenario and compare them against a baseline case with full surface coverage to quantify the degradation in reconstruction quality. \autoref{tab:partial_cov} summarizes the changes in the reconstruction quality when airfoil coverage is reduced and \autoref{fig:partial_cov} displays qualitative reconstruction results across the different sensor coverage scenarios.

\renewcommand{\arraystretch}{1.0}
\begin{table}[h!]
\caption{Changes in the reconstruction quality for the partial coverage FRGT-S models with regard to the full coverage baseline.}
\centering
\begin{tabular}{@{}lcccc@{}}
\toprule
\textbf{Metric}  & \begin{minipage}{.15\textwidth}
      \includegraphics[width=\linewidth]{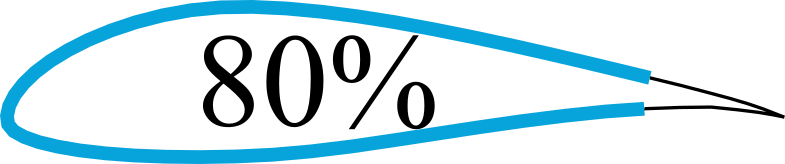}
    \end{minipage} & \begin{minipage}{.15\textwidth}
      \includegraphics[width=\linewidth]{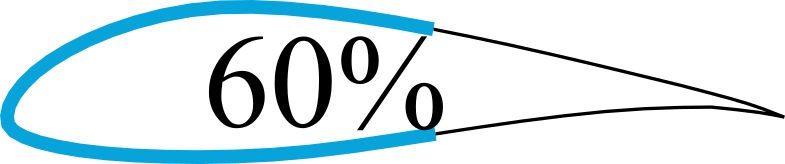}
    \end{minipage}  &\begin{minipage}{.15\textwidth}
      \includegraphics[width=\linewidth]{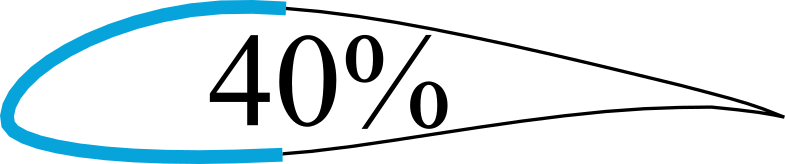}
    \end{minipage} &\begin{minipage}{.15\textwidth}
      \includegraphics[width=\linewidth]{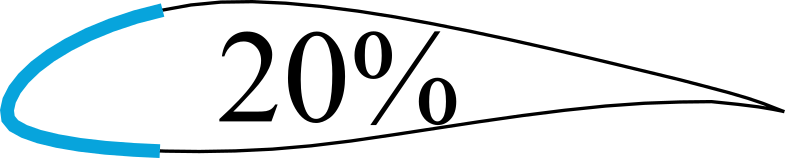}
    \end{minipage}\\ \midrule
Pressure RMSE change [\%] & +19.83 & +54.27& +72.22  & +142.44 \\
x-Velocity RMSE change [\%] & +10.02 & +22.18& +27.41 & +53.17 \\
y-Velocity RMSE change [\%] & +11.19 & +15.89& +16.93 & +29.87 \\
\bottomrule
\end{tabular}
\label{tab:partial_cov}
\end{table}

 The results reveal patterns in terms of how reduced sensor coverage affects the reconstruction of different flow variables. The pressure field exhibits the highest sensitivity to coverage reduction, with the RMSE increase reaching +142.44\% at 20\% coverage. This pronounced effect can be attributed to the high accuracy achieved with full coverage, where the baseline model already effectively captures the pressure patterns, thus leading to higher relative errors for reduced coverage. Velocity components demonstrate greater resilience to reduced coverage. The y-velocity component shows a particularly robust behavior, with only a +29.87\% RMSE increase even at 20\% coverage, while x-velocity maintains reasonable accuracy with a +53.17\% increase.

Qualitative analysis of the reconstructions reveals that fundamental flow physics is preserved even under limited sensor configurations. The model maintains coherent predictions of key features, including boundary layer structure and leading-edge pressure distribution, even down to 20\% coverage. However, the wake region exhibits increasing uncertainty as sensor coverage decreases, particularly evident in the pressure error distributions. 

These findings have important implications for practical MEMS-based sensing applications. Our results suggest that acceptable velocity reconstruction can be maintained with coverage as low as 20\%, though applications with stringent pressure accuracy requirements may need to balance between two approaches: either maintaining higher sensor density in critical regions, or compensating for reduced coverage through more data and/or more specialized training approaches. One could for instance consider a two-phase approach with a first model trained to reconstruct surface pressure followed by the FRGT for the full flow field.

\section{Discussion}
\label{sec:discussion}
The results presented in this work demonstrate the potential of hybrid Graph Transformer architectures for flow field reconstruction from sparse surface measurements on meshes. However, several key challenges and opportunities emerge from our findings that warrant further investigation. 

Firstly, while our FRGT architectures achieve promising results with approximately 1.4M parameters, this remains modest compared to modern large-scale deep learning models~\cite{touvron2023llama}. The primary constraint on scaling stems from the memory requirements of back-propagating gradients for our large mesh-based graphs, where even a single airfoil case can contain upwards of 50,000 nodes. The quadratic complexity of traditional attention mechanisms becomes prohibitive in this context, even with our adoption of linear attention variants. Scaling to much larger model capacities would likely require distributed training and more sophisticated memory management techniques.

Another challenge lies in the inherent structure of aerodynamic meshes. The necessarily high cell density near airfoil surfaces, critical for resolving boundary layer physics, creates an implicit optimization bias. The model naturally prioritizes accuracy in these dense near-wall regions, where the majority of nodes reside, potentially at the expense of far-field prediction quality. Future work might explore custom loss functions that better balance near- and far-field reconstruction accuracy, perhaps through adaptive weighting schemes based on mesh density. Moreover, our dataset reflects the traditional design philosophy of airfoils, which are often skewed for positive angle of attack operation. Our sampling strategy, which covers equally the full angle of attack range, results in an under-representation of separated flow conditions at positive angles, as separation typically occurs earlier at negative angles of attack for many airfoils. Future iterations could consider targeted data augmentation strategies to create a more balanced distribution of flow regimes. Additionally, while our use of unstructured triangular meshes provides good flexibility and interpolation opportunities, the consistent meshing strategy may allow the model to learn implicit patterns in the spatial discretization rather than the purely physical relationships. Although triangular elements offer natural adaptability, the model may develop biases towards expected cell arrangements and refinement patterns. Future work should investigate training on multiple discretization types, including structured meshes, hybrid topologies, and even random point clouds with Delaunay triangulation. This diversity would help ensure the model learns true flow physics rather than potentially exploiting specific mesh patterns.

Several promising directions exist for enhancing the FRGT architecture through physics-informed priors and uncertainty handling. A particularly compelling approach would be to combine our data-driven method with physics-based inflow estimation techniques. For instance, potential flow models coupled with conformal mapping can provide reliable estimates of key flow parameters like stagnation point location and freestream conditions from surface pressure measurements~\cite{marykovskiy2023hybrid}. These physics-derived estimates could serve as conditioning variables for our Graph Transformer, providing an informed prior that helps constrain the space of possible flow field reconstructions. Such physics-ML integration could be implemented through various strategies~\cite{haywood2024discussing}: regularization terms during training, embedding known conservation laws into the loss function, or even designing architecture components that explicitly respect physical symmetries. This could potentially improve reconstruction accuracy while also enhancing generalization to flow conditions outside the training distribution and reducing the amount of required training data.

Our partial coverage experiments reveal robustness in the FRGT's reconstruction capabilities. Even with only 20\% sensor coverage along the front half of the airfoil, the model maintains reasonable prediction accuracy, with velocity component errors increasing by only 53.17\% and 29.87\% for $u_x$ and $u_y$ respectively. Given the practical constraints of instrumentation in experimental aerodynamics, this robustness is particularly noteworthy. It is often difficult to achieve a comprehensive pressure tap coverage due to structural requirements, internal space limitations, or cost constraints. The demonstrated capability to reconstruct meaningful flow fields from limited sensor data opens up new possibilities for testing. For instance, rapid flow field estimates could enable real-time adjustment of wind tunnel parameters or flight test conditions, potentially reducing the number of test points required to characterize an aerodynamic configuration. Furthermore, the model's ability to handle partial measurements suggests applications in extended health monitoring, where progressive sensor failures could be compensated for without complete system replacement.

\section{Conclusion}
\label{sec:conclusion}
This work introduces the Flow Reconstruction Graph Transformer (FRGT), a hybrid architecture that combines message-passing neural networks with efficient Transformers to reconstruct aerodynamic flow fields from surface pressure measurements. We evaluate our models on a diverse dataset of airfoil configurations and demonstrate that this approach can effectively reconstruct both pressure and velocity fields while maintaining computational efficiency, with inference times around 200ms on consumer hardware.
Our results highlight several key findings. First, the combination of local geometric processing through message-passing and global information exchange via linear attention proves particularly effective for flow reconstruction tasks, outperforming pure message-passing approaches in both accuracy and computational efficiency. Second, the architecture demonstrates robustness to sensor coverage reduction, maintaining reasonable velocity field predictions even with only 20\% of the nominal sensor coverage. This resilience has important implications for practical sensing applications where comprehensive surface instrumentation may be infeasible.
The FRGT architecture also reveals interesting insights about the relative importance of local and global processing in mesh-based learning. Our ablation studies show that sufficient local geometric processing through message-passing is crucial before applying global attention mechanisms, suggesting that effective mesh-based learning requires careful balance between local structure preservation and long-range information exchange.
Looking forward, this work opens several promising directions for both theoretical development and practical applications. The demonstrated capability to handle partial measurements could enable new sensing and processing approaches to experimental aerodynamics and real-time monitoring. Additionally, the framework's ability to process arbitrary mesh topologies, while currently limited by consistent meshing strategies, provides a foundation for developing truly mesh-agnostic architectures. As computational resources continue to expand and techniques for efficient attention mechanisms evolve, scaling to larger model capacities could further enhance reconstruction accuracy, particularly for complex separated flows and unconventional geometries.
Ultimately, this research contributes to the growing body of work demonstrating the potential of geometric deep learning for physics problems. The success of the FRGT in reconstructing complex aerodynamic fields from limited surface data suggests similar approaches could be valuable across a broader range of physics-based inverse problems where state reconstruction from boundary measurements is required.

\bibliographystyle{apalike}
\bibliography{biblio}
\end{document}